\pdfoutput=1
\documentclass[11pt]{article}
\usepackage{amsmath}

\usepackage{ACL2023}

\usepackage{times}
\usepackage{latexsym}

\usepackage[T1]{fontenc}
\usepackage[utf8]{inputenc}

\usepackage{microtype}

\usepackage{inconsolata}

\usepackage{listings}
\title{A Baseline for Self-state Identification and Classification in Mental Health Data: CLPsych 2025 Task}

\author{Laerdon Kim \\
  Cornell University \\
  Ithaca, NY \\
  \texttt{lyk25@cornell.edu}}

\begin{document}
\maketitle
\begin{abstract}
    We present a baseline for the CLPsych 2025 A.1 task: classifying self-states in mental health data taken from Reddit. We use few-shot learning with a 4-bit quantized Gemma 2 9B model \cite{gemmateam2024gemma2improvingopen, brown2020languagemodelsfewshotlearners, unsloth} and a data preprocessing step which first identifies relevant sentences indicating self-state evidence, and then performs a binary classification to determine whether the sentence is evidence of an adaptive or maladaptive self-state. This system outperforms our other method which relies on an LLM to highlight spans of variable length independently. We attribute the performance of our model to the benefits of this sentence chunking step for two reasons: partitioning posts into sentences 1) broadly matches the granularity at which self-states were human-annotated and 2) simplifies the task for our language model to a binary classification problem. Our system places third out of fourteen systems submitted for Task A.1, achieving a test-time recall of 0.579.

    % what more can go here?
\end{abstract}

\section{Introduction}

Evaluating the mental state of a patient takes careful analysis of textual data. Large language models (LLMs) have demonstrated strong ability to comprehend intention, perception, and cognition conferred by natural language. This extends to mental health tasks; for example, CLPsych 2024 demonstrates the ability of LLMs to accurately capture fragments of evidence justifying the classification of suicide risk based on online Reddit posting \cite{chim-etal-2024-overview}. We seek to provide information on how simple LLM systems respond to different forms of data preprocessing to scaffold a complex task like that of classifying self-states. We explain two primary strategies we employ to boost performance on this self-state evidence identification and classification task: a preprocessing step using LLMs to identify "important" spans which provide information about the user's psychological state, and a system using an LLM to identify specific spans which evidence an adaptive or maladaptive self-state.

\section{Data}

The training data provided by the CLPsych 2025 organizers consists of 30 JSON files each containing a Reddit user's timeline, totaling 343 posts overall. Each timeline entry consists of two levels of structure: a timeline level, which contains a summary (string). Within each timeline is a post level, which contains one or more posts, each with a unique post ID (string). Each post includes four fields: adaptive evidence, maladaptive evidence, summary, and well-being score. The evidence fields contains a list of strings which correspond to substrings within the post text.

Evaluation of submissions for Task A.1 is recall-oriented: system performance is calculated using an average of the maximum pairwise BERTScore \cite{zhang2020bertscoreevaluatingtextgeneration}. For each predicted sentence, the highest BERTScore it achieves with any gold annotated sentence is taken, and these maximum scores are averaged. A secondary metric used to assess submissions is weighted recall, which recognizes systems that had a cumulative number of annotated tokens more similar to the number of human-annotated tokens \cite{tseriotou2025overview}.

\section{Methods}

For each of the methods described here, we use 4-bit quantized Gemma 2 9B, without fine-tuning. The prompts used to achieve these results are provided in Appendix A.

\subsection{Baseline}

To produce a baseline with our language model, we divide the post into sentences using spaCy \cite{honnibal2020spacy} and classify each sentence as adaptive or maladaptive. We provide definitions of adaptive and maladaptive self-states drawn from those provided in the task overview in our prompt.

\subsection{Context}

After initial runs of our baseline model, we sought to improve performance by providing the model with context (all previous sentences in the post) and using few-shot learning. We also add two examples of classification (one adaptive, one maladaptive) with a brief justification, and a detailed description of the MIND framework \cite{slonim2024self}.

\subsection{Importance filtering}

In order to increase the precision of our evidence extraction, we add a preprocessing step, using an LLM to first determine whether or not a sentence was "important" or not. We define this as containing some reference to any one of six MIND self-state dimensions--affective, behavior-self, behavior-others, cognition-self, cognition-others, and desire \cite{slonim2024self}.

\subsection{LLM span identification}

After analyzing low-recall posts, we notice that many self-state spans are annotated at a sub-sentence level. We find that 70.2\% of maladaptive and 68.7\% of adaptive self-states were not sentence spans (defined as starting with a capital letter and ending with punctuation). 23.6\% of adaptive spans and 19.2\% of maladaptive spans are <7 words long. For example, one maladaptive span begins with a comma, explains that medical professionals are unable to help them, and does not end with punctuation. One adaptive span simply states that nobody can be perfect, a brief sentence less than seven words long. 

In order to improve performance, we use the language model to identify self-states at a finer level, attempting to catch these sub-sentence spans. Our second method on Task A.1 separates the post into slightly larger contexts, and prompting our model to both identify and classify self-states independently. We use spaCy to again split sentences, and then merge them into 2-sentence groups. On each group, the model is then prompted to identify phrases at a sub-sentence level, and given the same information as the baseline in the prompt. The model returns a list of dictionaries containing substrings of the 2-sentence chunk and their predicted labels.

\subsection{LLM span identification with adaptive recall boost}

Low adaptive recall scores prompted us to experiment with explicitly steering the model to pay careful attention to subtle adaptive self-states embedded within sentences via prompting, noting that adaptive self-states may be hidden within seemingly maladaptive sentences, and encouraging the model to annotate as much of the chunk as possible. In addition, we modify the prompt to model this behavior in the examples, choosing larger substrings which collectively span over the entire chunk.

\section{Results}

\subsection{Baseline results}

Overall, our naive baseline method of classifying individual sentences of the post outperforms all methods except for Baseline + Context.

The addition of context provides a modest increase in adaptive self-state recall.

The addition of the importance filtering preprocessing step slightly degrades recall in exchange for improving weighted recall. On the sample training subset we use for evaluation in Table~\ref{tab:comparison}, the importance filtering reduces the number of spans considered from 370 to 232.

\subsection{LLM span identification (LLM Span ID) results}

Notably, our span identification system--which tasks our LLM with both identifying a self-state span and classifying it simultaneously--significantly increases maladaptive self-state recall by approximately 0.107 from the Baseline + Context + Importance, but also decreases adaptive self-state recall by 0.206.

Additionally, the LLM Span ID method offers weighted recall performance slightly below that achieved by importance filtering.

Our prompt steering in the LLM Span ID + Adaptive Boost row significantly improves adaptive recall by 0.107, at the expense of 0.033 points in maladaptive recall and losses in the weighted recall metric across both categories.

\par\bigskip
\begin{table*}[t]
\centering
\caption{Side-by-side comparison of methods, collected via a sample of five training set timelines. Overall vs.\ weighted metrics on separate rows. \textbf{A} indicates adaptive score; \textbf{M} indicates maladaptive score.}
\label{tab:comparison}
\begin{tabular}{lccc}
\hline
\textbf{Method} & \textbf{Overall Recall} & \textbf{Recall (A)} & \textbf{Recall (M)} \\
                    & \textbf{Weighted Recall} & \textbf{Weighted Recall (A)} & \textbf{Weighted Recall (M)} \\
\hline
Baseline & 0.504 & 0.452 & 0.556 \\
         & 0.196 & 0.204 & 0.188 \\
\hline
Baseline & \textbf{0.520} & \textbf{0.488} & 0.553 \\
(Context) & 0.201 & 0.204 & 0.197 \\
\hline
Baseline & 0.499 & 0.472 & 0.527 \\
(Context + Importance) & \textbf{0.244} & \textbf{0.266} & \textbf{0.221} \\
\hline
LLM Span ID & 0.450 & 0.266 & \textbf{0.634} \\
            & 0.220 & 0.248 & 0.193 \\
\hline
LLM Span ID & 0.487 & 0.373 & 0.601 \\
(Adaptive Boost) & 0.172 & 0.229 & 0.114 \\
\hline
\end{tabular}
\end{table*}
\par\bigskip

\section{Discussion}

\subsection{Challenges in capturing adaptive expressions}

The difficulty of capturing adaptive self-states is apparent in the results of Table~\ref{tab:comparison}. Adaptive recall consistently lags behind maladaptive recall. Observing human-annotated adaptive and maladaptive self-states reveals that adaptive self-states are generally much more subtle than their maladaptive counterparts. Annotations of adaptive self-states can contradict intuition; disappointment or anger can indeed signal an adaptive self-state if it reflects positive affective expression. Some adaptive self-states reference posters crying and breaking down into tears, or getting angry at others in their life like their partners. For an LM with limited context, it may be difficult to recognize such actions as adaptive signals.

For example, many adaptive self-states fall under categories of asking other posters for help, two-word interjections, or sentences describing the narrator's plans to do some common action, such as going to the store. A deep understanding of the poster's behavior overall is needed to assess whether or not the span signals adaptive thinking. In comparison, maladaptive self-states often reference self-harm, suicide, feelings of worthlessness--generally, these states contain some semantically similar terms.

In contrast, evidence reflecting maladaptive self-states is comparatively extreme, often explicitly referencing behaviors or perceptions ranging from self-harm to feelings of isolation.

While previous CLPsych tasks have demonstrated LLMs' strong performance in similar highlighting tasks with identifying evidence for suicide risk \cite{shing-etal-2018-expert}, the more subtle task of identifying patterns indicating psychological health is arguably more difficult \citep{zirikly-etal-2019-clpsych, tsakalidis-etal-2022-identifying}.

\subsection{Effects of context and importance filtering}
Providing context and filtering irrelevant sentences improved recall and weighted recall, respectively.

We hypothesize the model can identify healthy changes in behavior more accurately with the user's context of more negative behavior. As discussed in the previous subsection, many maladaptive evidence spans are more apparent than adaptive spans--as a result, adaptive recall benefited more from the added information.

Our second addition, importance filtering, improves weighted recall, but lowers overall recall.

While many sentences in the provided training data bear no relevance to self-states, even very subtle references to users' sense of self-worth and asking for help from other members of the subreddit can qualify as adaptive self-states, for example. Subtler spans of evidence from either category are likely wrongfully discarded during this step.

Notably, the importance filtering step does not have the post's context--it inferences sentence-by-sentence only. Sentences which may seem irrelevant without context can become important with context, which may explain the degradation of recall. 

Ultimately, we expect any filtering step to decrease recall to some extent.

\subsection{Baseline vs.\ LLM Span ID}

The superior performance of our original baseline system can be attributed to its more balanced recall across the two self-state categories. While the LLM Span ID strategy 
excels at identifying maladaptive self-states, it clearly fails to reliably identify adaptive evidence.

We propose that one potential reason for this divergence is the increased emotional intensity of maladaptive evidence spans compared to adaptive evidence spans. A manual qualitative analysis while randomly sampling pairs of adaptive and maladaptive spans corroborates our previous proposition that maladaptive evidence is generally more explicit. Many gold-standard adaptive spans contain more objective statements about the poster's situation, maladaptive spans generally reference negative events, like the loss of a friendship or falling into a depressive state, for example. Span-based prompting may encourage the model to be more conservative, favoring strictly unambiguous phrases as evidence.

On the other hand, our baseline removes guesswork from this partitioning subtask: sentence-level pre-slicing simplifies the model's task to a binary classification problem at a fixed granularity. This aligns well with the sentence-level granularity of human annotation, and reduces the cognitive burden on the model compared to identifying and labeling variable-length sub-sentence spans. If a sentence is classified as adaptive, the more coarse sentence-level classification may provide a better match to evidence spans which are also at about sentence length.

\section{Conclusion}

We present a simple approach to the highlighting task presented in CLPsych 2025, centering LLMs in our system and using primarily prompting and data processing strategies to maximize our performance. By comparing two methods, baseline sentence classification and LLM span identification, we demonstrate how some performance variance can be elicited simply by structuring a task differently.

We hope our work provides some insight into the behavior of large language models when grappling with complex emotional dimensions.

\section*{Limitations}

Our work has yet to explore a hybridized approach, potentially combining two distinct systems for adaptive and maladaptive classification. An adaptive span identification could be tuned to be more sensitive to subtler self-state dimension indications, whereas the maladaptive detection system could be designed more similarly to the high-performing vanilla LLM Span ID method. In addition, our preprocessing method of choosing 2-sentence long chunks for LLM span identification was not verified as an optimal choice--a 3-sentence sliding window may potentially be a better option, able to analyze each sentence in the context of sentences before and after it. 

\section*{Ethics}

This work was completed following the ACL code
of ethics. Each team member completed a data usage agreement form and received the password-protected dataset securely. Data used was uploaded to the secure Cornell Information Science compute cluster, removed immediately following completion of the task. Models used for inference were entirely open-source. We have paraphrased examples from the dataset and removed our examples from the prompts in the appendix.

\section*{Acknowledgements}

The author would like to thank the anonymous Reddit users whose posts are featured in this year's shared tasks, as well as the organizers of CLPsych at the Queen Mary University of London. In addition, the members of Team Zissou at Cornell University were invaluable to this submission: Dave Jung, Professor Cristian Danescu-Niculescu-Mizil, Professor Lillian Lee, Tushaar Gangavarapu, Sean Zhang, Son Tran, Vivian Nguyen, Luke Tao, Ethan Xia, and Yash Chatha.

\bibliographystyle{acl_natbib}
\bibliography{references}

\appendix

\section{Appendix}
\label{sec:appendix}

\subsection{Baseline}
\begin{lstlisting}
You are a professional psychologist. Given a social media post, classify whether or not a sentence demonstrates an adaptive or maladaptive self-state.
An adaptive self-state reflects aspects of the self that are flexible, non-ruminative, and promote well-being and optimal functioning.
A maladaptive self-state reflects internal states or perspectives that hinder an individual's ability to adapt to situations or cope with challenges effectively, potentially leading to emotional distress or behavioral problems.

Here is the sentence:
\end{lstlisting}

\subsection{Baseline (Context)}
\begin{lstlisting}
You are a professional psychologist. Given a social media post, classify whether or not a sentence demonstrates an adaptive or maladaptive self-state.

An adaptive self-state reflects internal processes that are flexible, constructive, and promote emotional well-being, effective functioning, and psychological health.
A maladaptive self-state reflects internal processes that are rigid, ruminative, self-defeating, or harmful, and are often associated with emotional distress or impaired functioning.

To make your classification, use the ABCD framework for psychological self-states:

A. **Affect** -- Type of emotional expression
   - Adaptive: calm, content, assertive, proud, justifiable pain/grief
   - Maladaptive: anxious, hopeless, apathetic, aggressive, ashamed, depressed

B. **Behavior** -- Main behavioral tendencies
   - Toward Others (BO):
     - Adaptive: relational, autonomous behavior
     - Maladaptive: fight/flight response, controlling or overcontrolled behavior
   - Toward Self (BS):
     - Adaptive: self-care
     - Maladaptive: self-neglect, avoidance, self-harm

C. **Cognition** -- Main thought patterns
   - Toward Others (CO):
     - Adaptive: perceiving others as supportive or related
     - Maladaptive: perceiving others as detached, overattached, or autonomy-blocking
   - Toward Self (CS):
     - Adaptive: self-compassion and acceptance
     - Maladaptive: self-criticism

D. **Desire** -- Expressed needs, goals, intentions, or fears
   - Adaptive: desire for autonomy, relatedness, self-esteem, care
   - Maladaptive: fear that these needs won't be met

Here are a couple of examples:
"--removed--"
This is maladaptive. It shows a bluntedness and apathic affective state.

"--removed--"
This is adaptive. The crying is not a sign of maladaptive self-state, rather it is a healthy sadness.

You will be shown:
1. The context of the post so far
2. The current sentence to classify

If the sentence clearly demonstrates one or more **maladaptive or adaptive self-state(s)** based on this framework, classify it accordingly.

Here is the post so far:
{context}

Here is the current sentence:
{sentence}
\end{lstlisting}

\end{document}